\documentclass[11pt, oneside]{article}
\usepackage{geometry}                		
\usepackage[utf8]{inputenc} 
\usepackage[T1]{fontenc}    
\usepackage{hyperref}       
\usepackage{url}            
\usepackage{booktabs}       
\usepackage{amsfonts}       
\usepackage{nicefrac}       
\usepackage{microtype}      
\usepackage{cite}
\usepackage[tableposition=top]{caption}
\usepackage{times}
\usepackage{graphicx} 
\usepackage{subfigure} 
\usepackage{multirow}
\usepackage{xcolor}
\usepackage{outlines}

\usepackage[utf8]{inputenc}
\usepackage[T1]{fontenc}
\usepackage{amsmath}
\usepackage{algorithm}
\usepackage{algpseudocode}
\newcommand{\varA}[1]{{\operatorname{#1}}}
\usepackage{multirow}
\usepackage{caption}
\usepackage{float}

\usepackage{hyperref}

\title{Robust registration of medical images in the presence of spatially-varying noise}

\author{Reza Abbasi-Asl\footnote{Corresponding author}\\
Department of Neurology, University of California, San Francisco, CA\\
Weill Institute for Neurosciences, University of California, San Francisco, CA\\
\vspace{10pt}
\textit{Reza.AbbasiAsl@ucsf.edu}\\
Aboozar Ghaffari \\
Electrical Engineering Department,\\
\vspace{10pt}
Iran Science and Technology University, Tehran, Iran\\
Emad Fatemizadeh \\
Electrical Engineering Department,\\
Sharif University of Technology, Tehran, Iran}

\date{}
\begin{document}

\maketitle


\begin{abstract}
Spatially-varying intensity noise is a common source of distortion in medical images. Bias field noise is one example of such a distortion that is often present in the magnetic resonance (MR) images or other modalities such as retina images. In this paper, we first show that the bias field noise can be considerably reduced using Empirical Mode Decomposition (EMD) technique. EMD is a multi-resolution tool that decomposes a signal into several principle patterns and residual components. We show that the spatially-varying noise is highly expressed in the residual component of the EMD and could be filtered out. Then, we propose two hierarchical multi-resolution EMD-based algorithms for robust registration of images in the presence of spatially varying noise. One algorithm (LR-EMD) is based on registration of EMD feature-maps from both floating and reference images in various resolution levels. In the second algorithm (AFR-EMD), we first extract an average feature-map based on EMD from both floating and reference images. Then, we use a simple hierarchical multi-resolution algorithm to register the average feature-maps. For the brain MR images, both algorithms achieve lower error rate and higher convergence percentage compared to the intensity-based hierarchical registration. Specifically, using mutual information as the similarity measure, AFR-EMD achieves 42\% lower error rate in intensity and 52\% lower error rate in transformation compared to intensity-based hierarchical registration. For LR-EMD, the error rate is 32\% lower for the intensity and 41\% lower for the transformation. Furthermore, we demonstrate that our proposed algorithms improve the registration of retina images in the presence of spatially varying noise.
\end{abstract}




\section{Introduction}

Accurate registration and alignment of two images has been a challenging problem in a wide variety of applications such as medical image processing \cite{sotiras2013deformable,feng2016liver}, remote sensing \cite{brooks1998multi}, biology \cite{wang2015fully,wang2014robust,matsuda2018accurate}, and computer vision \cite{lucas1981iterative,wang2017augmented}. Particularly, registration of the medical images has been widely used in tumor localization and targeting \cite{brock2006feasibility}, organ growth studies \cite{thompson2000growth} and brain atlas reconstruction \cite{vemuri2003image}. Overall, the registration algorithms can be categorized into two classes \cite{oliveira2014medical}. One class is concerned with the registration based on the intensity of the images. In this approach, a similarity measure is defined to quantify the similarity of both floating (or moving) and reference (or target) images. Then an optimization process identifies an optimal map for the floating image to achieve the highest similarity to the reference image. In the second class of registration algorithms, a set of features such as landmarks \cite{johnson2002consistent}, histogram of intensity \cite{ghanbari2012automatic} or responses to Gabor \cite{shen2002hammer} and Alpha stable filters  \cite{liao2010feature, abbasi2011mmro} are first extracted from both floating and reference images. Then two images are aligned through the mapping of these features. 

A similarity measure between two images is a crucial component of the intensity-based image registration. A wide variety of measures such as sum-of-squared-differences (SSD), correlation coefficient (CC) or mutual information (MI) \cite{wells1996multi, maes1997multimodality, guyader2018groupwise} have been used in the registration process. However, none of these measures are robust against the spatially varying intensity distortion. The bias field noise is one of the common instances of the spatially varying distortion \cite{studholme2006deformation} and is often present in magnetic resonance (MR) or retina images. Therefore the traditional registration algorithms may easily fail for the MR or retina images when these images are distorted.

Traditionally, two distinct types of algorithms have been studied to reduce the bias-field noise effect in image registration. One type is focused on defining robust similarity measures. Residual complexity (RC) \cite{myronenko2010intensity}, rank-induced \cite{ghaffari2015rism, ghaffari2017image} and sparsity-based \cite{afzali2016medical, ghaffari2015robust} similarity measures are some examples. The second type of algorithms are based on simply reducing the noise effect and then registration of the denoised images \cite{pohl2005unifying, friston1995spatial}. Our two proposed algorithms in this paper could be classified in the second group. We examine the ability of a state-of-the-art hierarchical signal decomposition techniques in removing spatially varying noise in MR images. Specifically, we propose to use empirical mode decomposition (EMD) \cite{huang1998empirical} to overcome the effect of bias field noise in MR images. EMD is a multi-resolution computational technique to decompose the image into several principle patterns and residual components. We show that the bias field noise is highly expressed in the residual components and not in the principle patterns. Therefore, it is possible to remove the noise by removing the EMD residuals. EMD-based algorithms have been previously studied for the basic image registration task \cite{riffi2013medical, jinsha2009multimodal}. However, to our knowledge, the advantage of the EMD-based registration algorithms in the presence of spatially varying noise is still unknown. In this study, we explore this advantage through proposing robust EMD-based registration algorithms. 

Our contributions in this paper are two folds: First, we show that the EMD is able to extracts the bias field noise from the image through multi-scale decomposition of the image. Second, we introduce two multi-resolution \cite{looney2009multiscale}  EMD-based registration algorithms robust to the bias field noise. In one algorithm (LR-EMD), we hierarchically register EMD feature-maps from both floating and reference images. The feature maps are constructed from various resolutions of EMD-based principle patterns. In the second algorithm (AFR-EMD), we first extract an average feature-map for both floating and reference images by averaging the principle patters from the EMD. Then, we use a simple hierarchical multi-resolution algorithm based on downsampling to register the average feature-maps. 
 
The rest of the paper is organized as the followings: In section 2, we describe the basis of signal decomposition using EMD. The application of EMD in extracting the bias field noise along with two EMD-based registration algorithms are discussed in section 3. The quantitative and qualitative registration performance and comparisons to the benchmark algorithms has been explored in section 4 for both brain MR and retina images. The study is concluded in section 5.

\section{Empirical mode decomposition}

Decomposing nonlinear and non-stationary signals into their intrinsic modes has been a challenging task in signal processing. Most of the previously proposed decomposition are based on time-frequency transformations such as Short-Time Fourier Transform \cite{schafer1973digital} and wavelet transform \cite{antonini1992image} that expands the signal into a set of basis functions. Unlike these transformations, algorithms based on empirical mode decomposition (EMD) \cite{huang1998empirical, wu2004study} are able to decompose the linear or nonlinear signal into a set of functions defined by the signal itself \cite{huang2003applications}. EMD consists of a set of spatial and temporal processes that decomposes signal into Intrinsic Mode Functions (IMFs). Each IMF is an oscillating mode of the signal. In contrast with harmonic modes, oscillating modes have time-variant amplitude and frequency. EMD has been widely used in applications of signal decomposition such as prediction \cite{ren2015comparative}, classification \cite{park2013classification} and denoising \cite{hao2017joint}. 

Bidimensional empirical mode decomposition \cite{nunes2003image} is particularly our focus in this paper, since it deals with two-dimensional signals such as images. Each IMF in the bidimensional empirical mode decomposition could be extracted from the image using a sifting process. Let $\mathbb{I}$ denote the image, $\mathbb{IMF}_i$ and $\mathbb{RES}_i$ denote the $i^{th}$ IMF and residual, respectively. Here $i$ represents the level or scale of the decomposition. For simplicity, let's assume that the residual at level 0 is the original image. That is:

$$\mathbb{RES}_0=\mathbb{I}.$$ 

To find the IMF in level $i$, we iterate through the following steps: First, identify all the local minimum and maximum of the $\mathbb{RES}_{i-1}$. Then, generate the bounding maximum and minimum envelopes, $\mathbb{E}^{max}$ and $\mathbb{E}^{min}$ using an interpolation or a surface fitting techniques. The IMF in this iteration is equal to $\mathbb{RES}_{i-1}$ subtracted by the mean of maximum and minimum envelope:

$$\mathbb{IMF}_i = \mathbb{RES}_{i-1} - \frac{\mathbb{E}^{max}+\mathbb{E}^{min}}{2}$$

We repeat this process until a convergence condition is met. At this point, the residual is defined as:

$$ \mathbb{RES}_{i} = \mathbb{RES}_{i-1} - \mathbb{IMF}_i $$.

The first IMF has the finest and the most detailed patterns while the rest of the IMFs have patterns with coarser content. The extracted IMFs have a number of crucial characteristics. First, the number of zero-crossing points is equal or one less than the number of extremas in each IMF. Second, each IMF has one minimum with zero value. Finally, the mean value of the bounding maximum and minimum envelopes is equal to zero for each IMF. 

It is known that the EMD has limited capacity in processing narrow-band signals \cite{chen2003technique}. These signals have concentrated energy in a narrow frequency band which causes the decomposition to fail in extracting the intrinsic modes. Thus, the accuracy and efficiency of the EMD could be limited in this situation. However, this is not an issue in our study for MRI and retina images. These images do not have narrow-band frequency component and therefore, are suited to be analyzed through EMD.
 

\begin{figure}[!t]
   \centering
   \includegraphics[width=.4\textwidth]{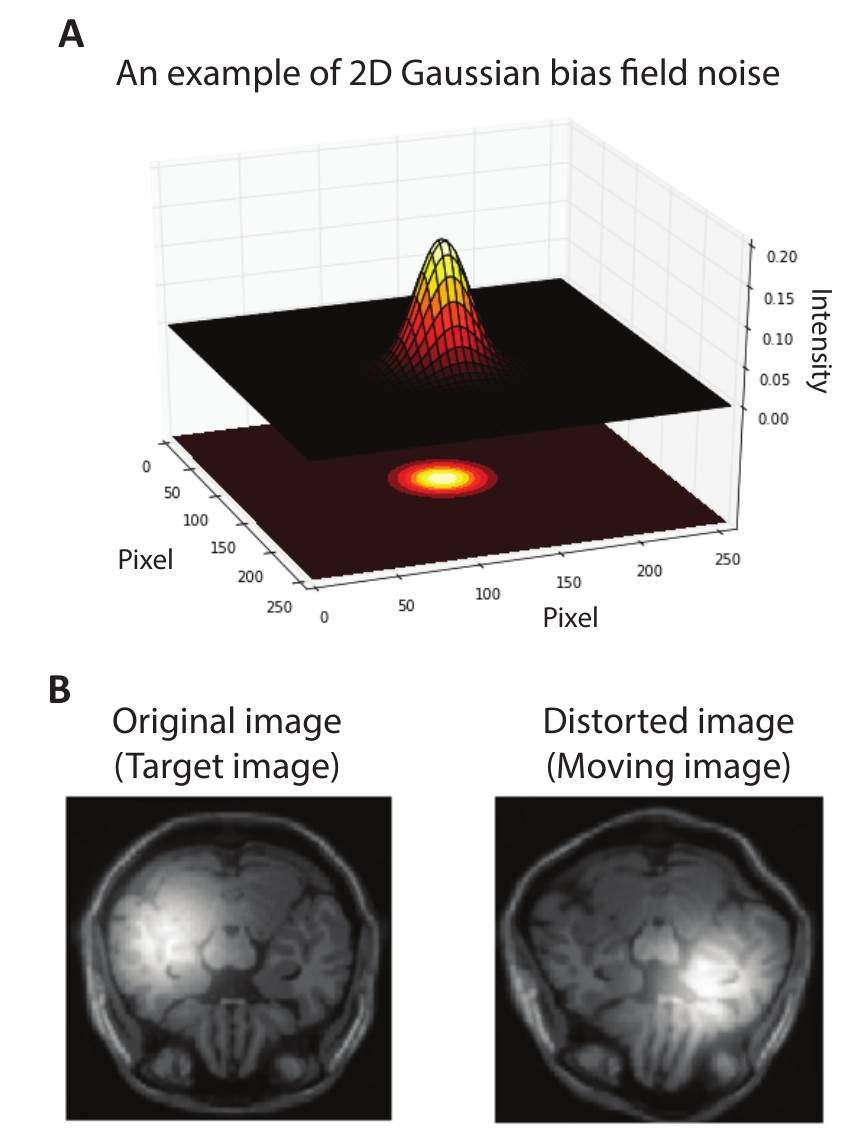}
   \caption{Simulated bias field noise in MR image. A. Sample two-dimensional Gaussian function. B. We manually add A Gaussian function with random mean to both floating and reference MR images.}
   \label{fig:biasfield}
\end{figure}

\section{Bias field noise and empirical mode decomposition}

Spatially varying noise or bias field is a common source of noise in MR images. Bias field noise can potentially cause the registration process to diverge, particularly, when similarity measure is not robust to the noise. Sum-of-squared-differences (SSD), correlation coefficient (CC) and mutual information (MI) are some of the non-robust similarity measures. Our results suggest that even Residual Complexity (RC) similarity measure has limited capacity in overcoming the bias field noise. In this section, we investigate the ability of EMD in reducing and removing this noise during the registration process. 

EMD is a powerful tool to extract the biad field noise from the image. Here, we provide an intuitive explanation for this statement. EMD provides a decomposition of the signal into its principle components in different levels of granularity. The principle patterns or the IMFs in each level (scale) contain a portion of the signal with a particular level of detail. The remaining components of the signal are explained in the residual part. Defined by the sifting process, each IMF is constructed by iteratively subtracting the average of the minimum and maximum envelopes from the original signal. In a two-dimensional space, the bias field noise could be estimated as a local spatial variation in the signal. The average of the minimum and the maximum envelopes is highly affected by this local spatial variation because both signals carry low-frequency patterns. Therefore the the average of the minimum and the maximum envelopes absorbs the bias field noise. By subtracting this average envelope from the signal, IMF becomes noise free.

\begin{figure}[!t]
   \centering
   \includegraphics[width=.5\textwidth]{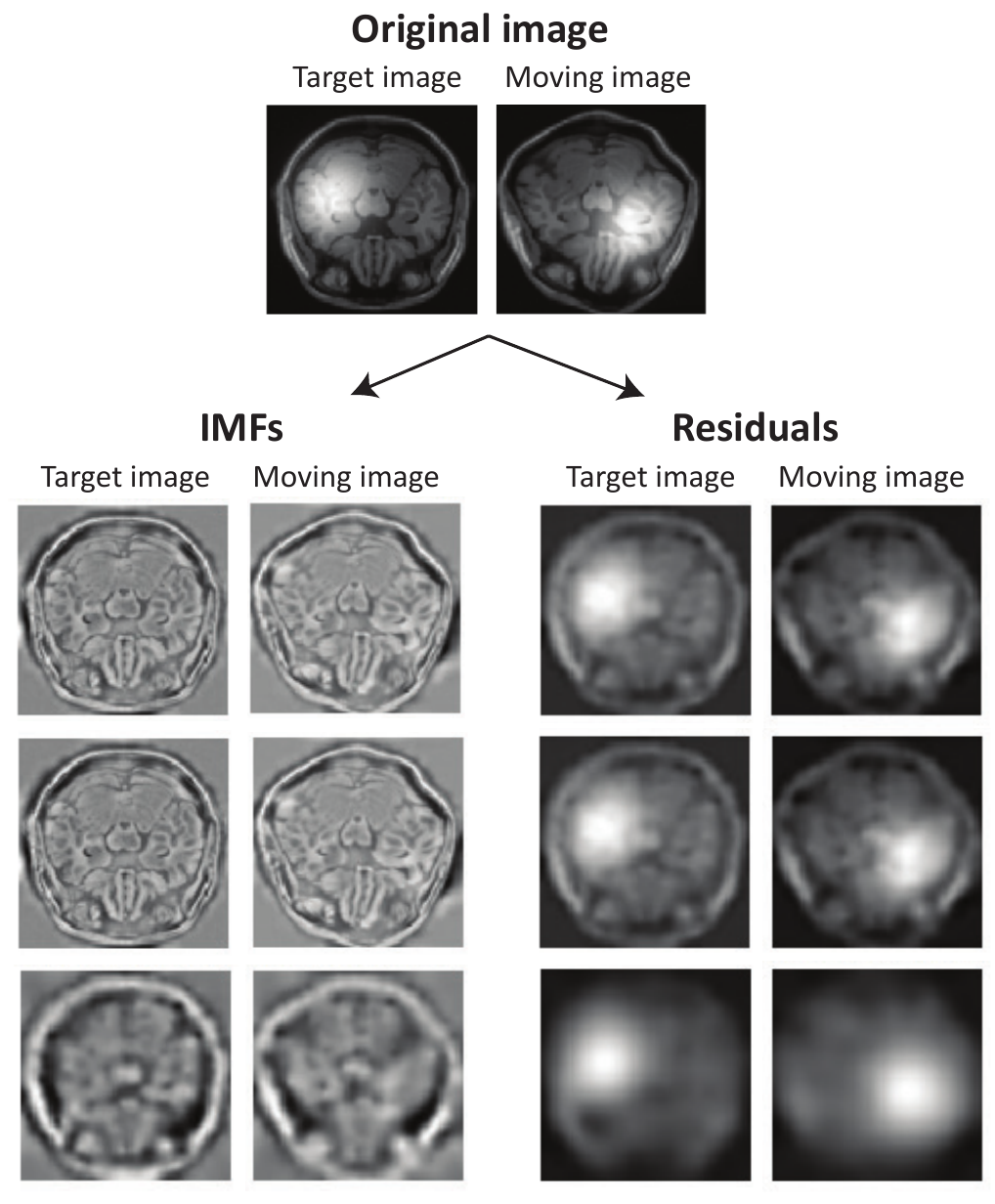}
   \caption{EMD separates bias field noise from the intrinsic components. The set of IMFs and residuals for three levels of EMD are shown for both floating and reference images. The Gaussian function is explained in the residuals and IMFs are noise free.}
   \label{fig:emd}
\end{figure}

Mathematically, the bias field noise can be represented by mixture of two-dimensional Gaussian functions (kernels) \cite{myronenko2010intensity}. We distort the image $I(x,y)$ with bias field noise according to the following equation:
\begin{equation}
I_{distorted}(x,y) = I(x,y) + \frac{1}{K}\sum_{k=1}^K e^{-||([x,y]-\mu_k||^2)/2\sigma^2}
\label{gaussian}
\end{equation}

where $K$ is the number of Gaussian functions. Figure \ref{fig:biasfield}-A shows a sample Gaussian kernel. We manually add one Gaussian kernel ($K=1$) to both floating and reference MR images. The coordination of center of Gaussion kernel is selcted at random and the standard deviation is set to be $\frac{1}{16}$ of the image width. \ref{fig:biasfield}-B shows a sample MR image from the BrainWeb dataset \cite{BrainWeb} corrupted with the bias field noise. To explore the ability of the EMD in extracting the bias field noise, we decomposed both noisy floating and reference images into IMFs and residuals. Figure \ref{fig:emd} illustrates the set of IMFs and residuals for three EMD levels. The Gaussian kernel is highly expressed in residuals and IMFs are noise free. This observation suggests that the registration algorithms based on the IMFs may be robust to the bias field noise. In the following two subsections, we propose two registration algorithms based on this observation.

\subsection{LR-EMD: Image registration algorithm based on EMD levels}

Taking advantage of the EMD's capacity in removing the bias field noise, we propose the following hierarchical multi-resolution registration algorithm. First, we find the IMFs for both floating and reference images. These IMFs represent the feature-maps in various resolutions. Starting from the coarsest resolution, we estimate the transformation from floating image IMF to the reference image IMF via free form deformation (FFD) algorithm \cite{rueckert1999nonrigid}. The estimated transformation provides an initial mapping for the free form deformation in the next level with the higher resolution. The process continues until registration of the finest level IMFs. We call this algorithm \emph{Level-based Registration using EMD or LR-EMD}. Note that other tranformation techniques could be used instead of the free form deformation. Our algorithm is also agnostic to the type of the similarity measure. In this paper, we have used SSD, CC, RC and MI as similarity measures. Algorithm \ref{alg:hier} shows the details of LR-EMD.   

LR-EMD takes advantage of IMFs in different levels (scales) and performs registration in a hierarchical order. Therefore, it benefits from the information stored in each level of IMFs during the process of the registration. The registration at each resolution level of IMFs provides an initial transformation for the registration in the next level. In principle, the hierarchical order increases the robustness of LR-EMD against local minimum in the optimization process.

\begin{algorithm}[tb]
		   \caption{LR-EMD: Image Registration using EMD levels}
		   \label{alg:hier}
		\begin{algorithmic}
		   \State {\bfseries Input:} $I_{fl}$ floating image, $I_{ref}$, reference image, $n$ number of IMFs or EMD levels, $\mathbb{SIM}$ Similarity measure
		   \State Extract $n$ IMFs of both $I_{fl}$ and $I_{ref}$,  
           \State Initialize registration with unity transform $f(x) = x$
		   \For{$i=1$ {\bfseries to}  $n$}
           \Comment where $i=1$ is the coarse-grained level and $i=n$ is the fine-grained level
		   \State Register the $i$th IMF of $I_{fl}$ to the $i$th IMF of $I_{ref}$ based on $\mathbb{SIM}$ Similarity measure and find transform $f(x)$ 
           \State Initialize transform for next level of IMF with $f(x)$ 
		   \EndFor
		\end{algorithmic}
\end{algorithm}

\subsection{AFR-EMD: Image registration algorithm based on average EMD feature-maps}

We propose a second registration algorithm based on the average of EMD-based features. This algorithm is called \emph{Average Feature-map Registration using EMD or AFR-EMD}. Algorithm \ref{alg:single} presents the details of AFR-EMD. In this case, we first construct a single feature-map for each of the floating and reference images by taking average of IMFs from all EMD levels (scales). Then, we register the average feature-maps based on the chosen similarity measure. To take advantage of hierarchical registration, we downsample the feature-map and perform a level-by-level registration. Similar to LR-EMD, various similarity measures including SSD, CC, RC and MI are used to perform AFR-EMD. Free form deformation \cite{rueckert1999nonrigid} is used here to estimate the transform.

The average of the IMFs is a denoised and normalized feature-map. Averaging the IMFs provides a feature map with reduced size. Therefore, the features are easier to interpret for a human observer. This reduction in the feature size could also reduce the computational cost in a single-scale registration algorithm. In our case, we designed a multi-scale hierarchical registration based on downsampling in AFR-EMD, therefore the gain in computational cost is in the same order as LR-EMD.  In the next section, we will explore the ability of LR-EMD and AFR-EMD in robust registration of images in the presence of bias field noise.




\section{Results}

We evaluated our proposed registration algorithms on the magnetic resonance (MR) images available from the BrainWeb dataset \cite{cocosco1997brainweb, BrainWeb}. Both algorithms were also evaluated for the registration of retina images. For the MR Images, the BrainWeb dataset contains simulated brain MR volumes from several protocols, including T1-weighted (MR-T1), T2-weighted (MR-T2), and proton density (MR-PD). Here, we selected a two-dimensional slice of the MR-T1 ($218 \times 181$). The intensities were normalized between 0 and 1.


Two different measures were used in this paper to compare the accuracy of registration. First measure is the transformation root mean square error (RMSE) between the true and estimated transformations. We call this measure T-RMSE throughout this paper. T-RMSE is equal to:

$$ \varA{T-RMSE} = \sqrt[]{\frac{1}{N}||T_{true} - T_{estimated}||^2} $$

Where $N$ is the number of pixels. Second measure is RMSE between intensities of the reference and the registered images. We call this measure I-RMSE. I-RMSE is equal to:

$$ \varA{I-RMSE} = \sqrt[]{\frac{1}{N}||I_{reference} - I_{registered}||^2} $$

The mean and the standard deviation (SD) of both error measures were calculated across 15 random initialization of the registration algorithms. At each run, the intensity distortion and the geometric transform parameters were randomly reinitialized.

\begin{algorithm}[tb]
		   \caption{AFR-EMD: Image registration based on Average EMD feature-maps}
		   \label{alg:single}
		\begin{algorithmic}
		   \State {\bfseries Input:} $I_{fl}$ floating image, $I_{ref}$, reference image, $n$ number of IMFs or EMD levels, $\mathbb{SIM}$ Similarity measure
		   \State Extract $n$ IMFs for both $I_{fl}$ and $I_{ref}$,  
           \State Find $\mathbb{F}_{fl}$ and $\mathbb{F}_{ref}$, the average of IMFs for both $I_{fl}$ and $I_{ref}$
           \State Initialize registration with unity transform $f(x) = x$
		   \For{$i=1$ {\bfseries to}  $n$}
           \Comment where $i=1$ is the coarse-grained level and $i=n$ is the fine-grained level
           \State Compute $\mathbb{F}_{fl}^i$ and $\mathbb{F}_{ref}^i$, the downsampled versions of $I_{fl}$ and $I_{ref}$ with respect to scale level $i$
		   \State Register $\mathbb{F}_{fl}^i$ to $\mathbb{F}_{ref}^i$ based on $\mathbb{SIM}$ Similarity measure and find transform $f(x)$ 
           \State Initialize transform for next level of registration with $f(x)$ 
		   \EndFor
		\end{algorithmic}
\end{algorithm}  

\subsection{BrainWeb dataset}

We used the simulated MR images in the BrainWeb dataset to evaluate the performance of our registration algorithms. Each image was geometrically distorted using a random perturbation to generate the floating image. Specifically, a $14 \times 14$ uniform grid was randomly perturbed (by a uniform distribution on [-6 6]) and used as the Free Form Deformation (FFD) grid for the floating image. Then, the distorted images were registered back to the original image using our two algorithms as well as the benchmark intensity based algorithm. For each algorithm, four registration processes were performed, each using either one of the SSD, CC, RC or MI similarity measures. In all of the cases, FFD \cite{rueckert1999nonrigid} was used as the model of geometric transform. The FFD, was implemented via three hierarchical levels of B-spline control points. We set the number of levels in IMFs, $n$, to 3 in order to have a fair comparison with the intensity-based registration. Iterative Gradient descent was used to optimize the transformation parameters. The accuracy was evaluated using both T-RMSE and I-RMSE.

We started the experiments by registering images without any bias field noise (section 4.1.1). Then, we manually added spatially varying distortions or bias field noise to both floating and reference images \cite{myronenko2010intensity}. The noise was generated using the mixture of Gaussian functions and was added to the images according to the Equation \ref{gaussian}. The mean of the Gaussian functions were selected at random. The standard deviation was set to $\frac{1}{16}$ of the image width.

\begin{figure}[!t]
  \centering
  \includegraphics[width=.5\textwidth]{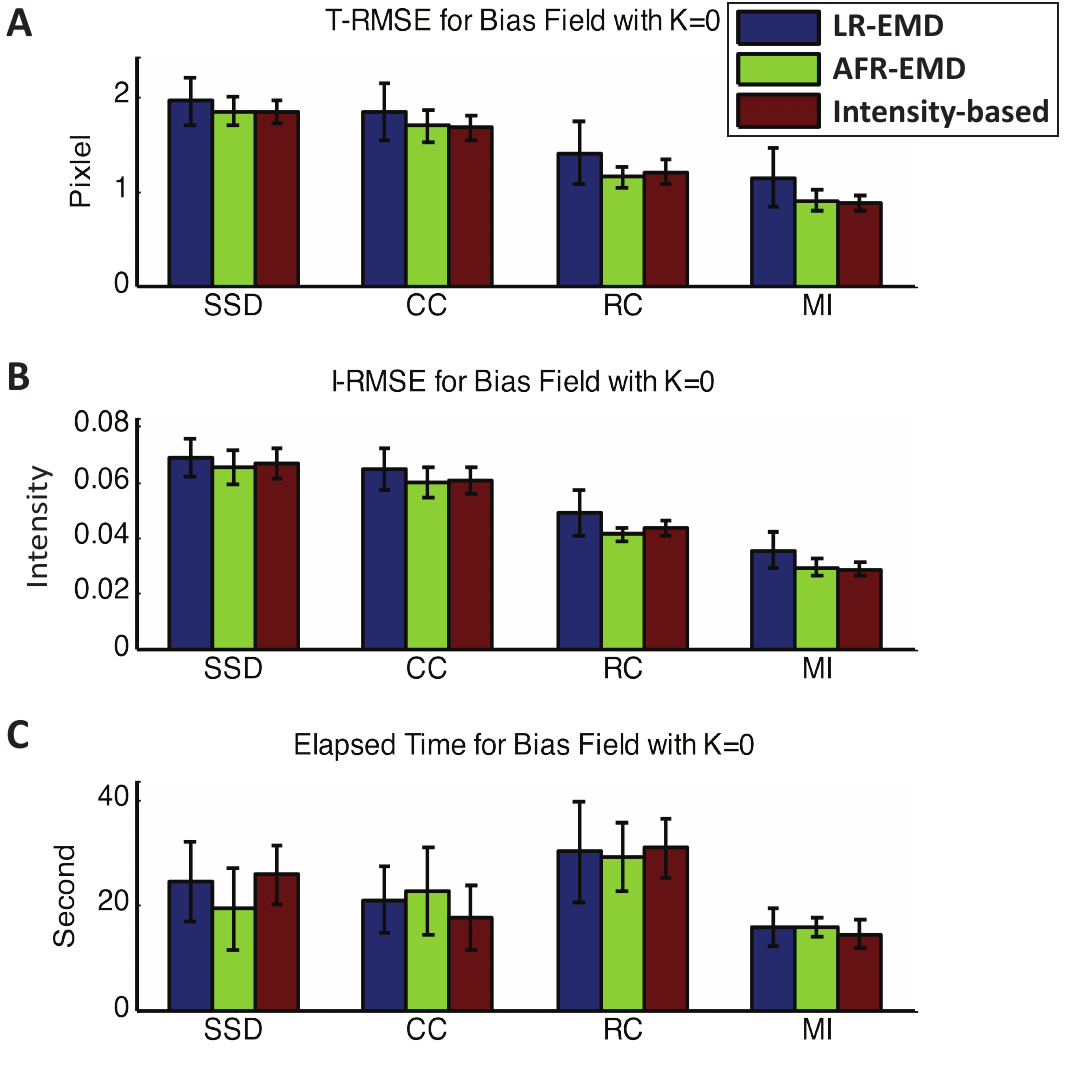}
  \caption{Registration performance when there is no bias field. Mean and variance of T-RMSE and I-RMSE for images in BrainWeb dataset are shown in top and middle panels. Bottom panel compares the running time for all methods.}
  \label{fig:4-10}
\end{figure}

\subsubsection{No bias field case}

Figure \ref{fig:4-10} shows the registration performance when there is no bias field in the MR images. The mean and variance of T-RMSE and I-RMSE for either of the SSD, CC, RC, and MI similarity measures are shown separately in the top and the middle panels. The performance metrics are reported for LR-EMD, AFR-EMD and the benchmark model. Without the bias field noise, all the registration processes for all images have converged. AFR-EMD achieves a similar performance to the benchmark intensity-based registration when there is no bias filed noise. LR-EMD has slightly higher error, particularly when RC has been used as the similarity measure. Both SSD and CC have approximately 25\% higher average error compared to the RC and MI similarity measures (both in T-RMSE and I-RMSE). Generally, MI achieves the highest accuracy and fastest convergence time among the similarity measures. The bottom panel in Figure \ref{fig:4-10} compares the running times for all the methods. There is no significant difference between running times of our proposed methods and the benchmark method. Among the similarity measures, RC has the highest running time. Overall, when there is no bias field distortion, EMD-based registration algorithms perform similar to the benchmark intensity-based method.

\begin{figure}[!t]
  \centering
  \includegraphics[width=.5\textwidth]{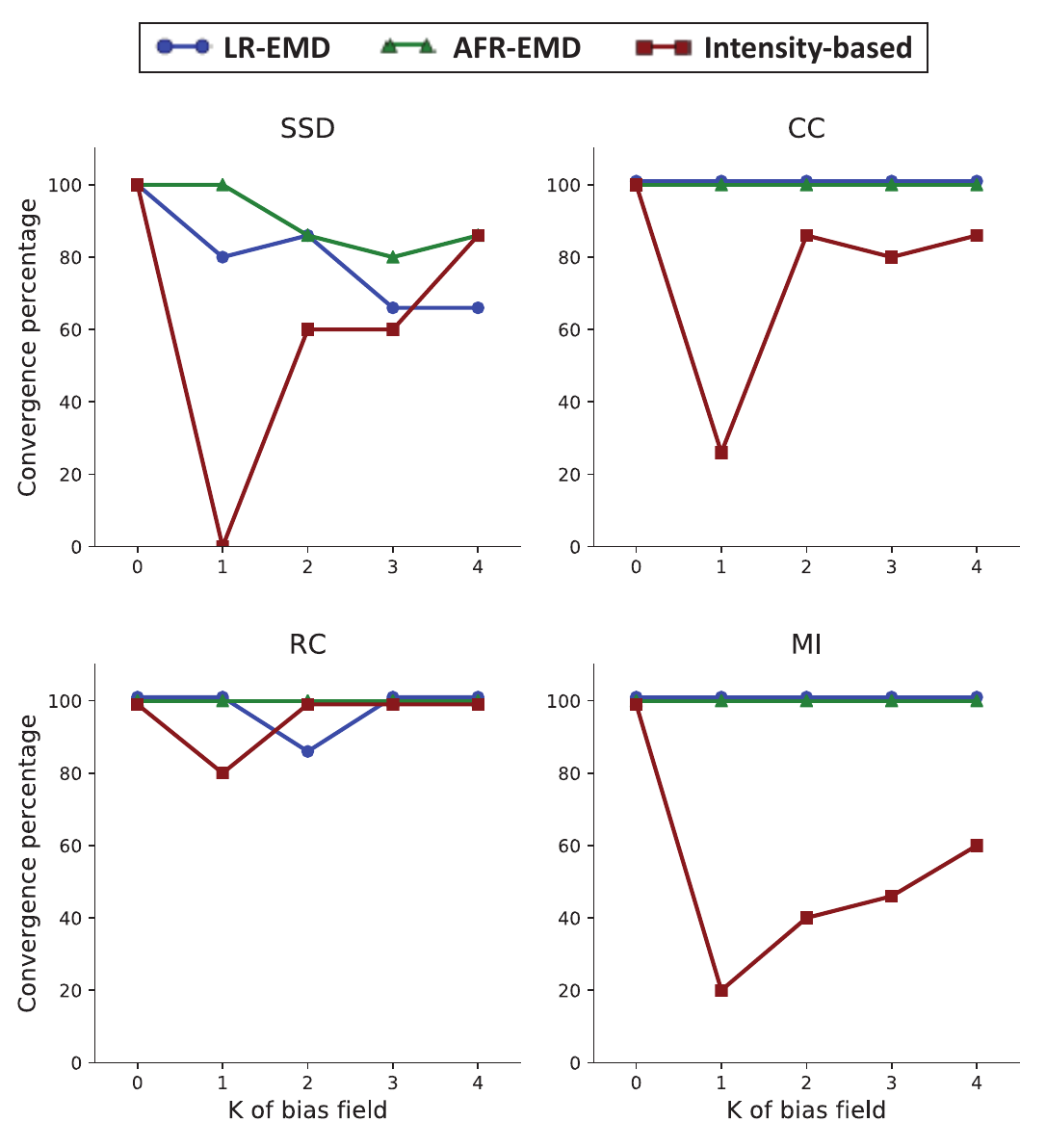}
  \caption{The convergence rate for baseline hierarchical registration with downsampling and our proposed algorithms based on EMD. Each panel corresponds to one similarity measure (SSD, CC, RC and MI). Here, the convergence is defined as T-RMSE error lower than 4 pixels.}
  \label{fig:4-11}
\end{figure}

\begin{figure*}[t]
\centering
\begin{minipage}{.48\textwidth}
  \centering
  \includegraphics[width=1\textwidth]{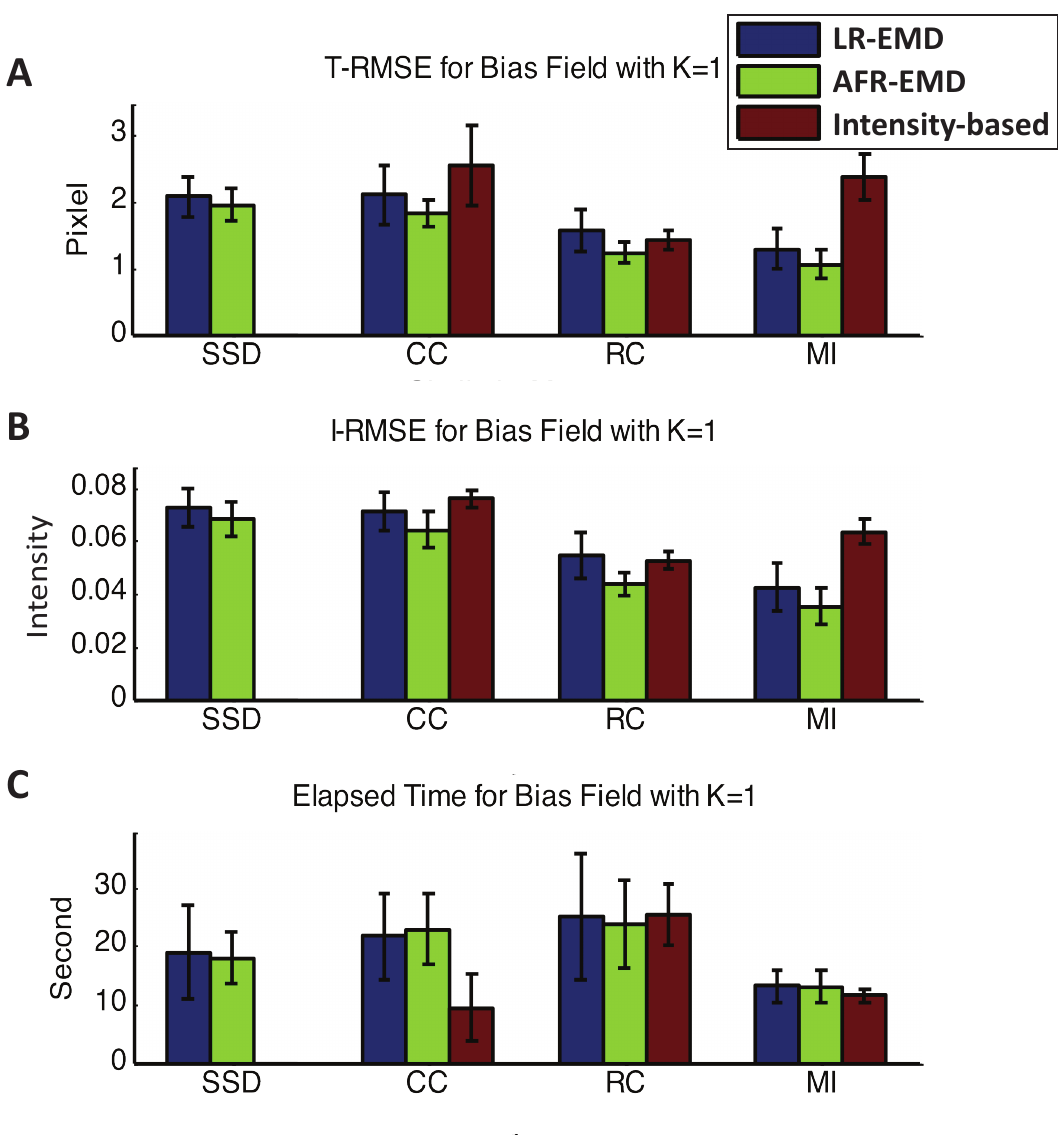}
  \captionof{figure}{Registration performance in the presence of one Gaussian kernel. Mean and variance of T-RMSE and I-RMSE for images in BrainWeb dataset are shown in top and middle panels. Bottom panel compares the running time for all methods.}
  \label{fig:4-15}
\end{minipage}%
\hfill
\begin{minipage}{.48\textwidth}
  \centering
  \includegraphics[width=1\textwidth]{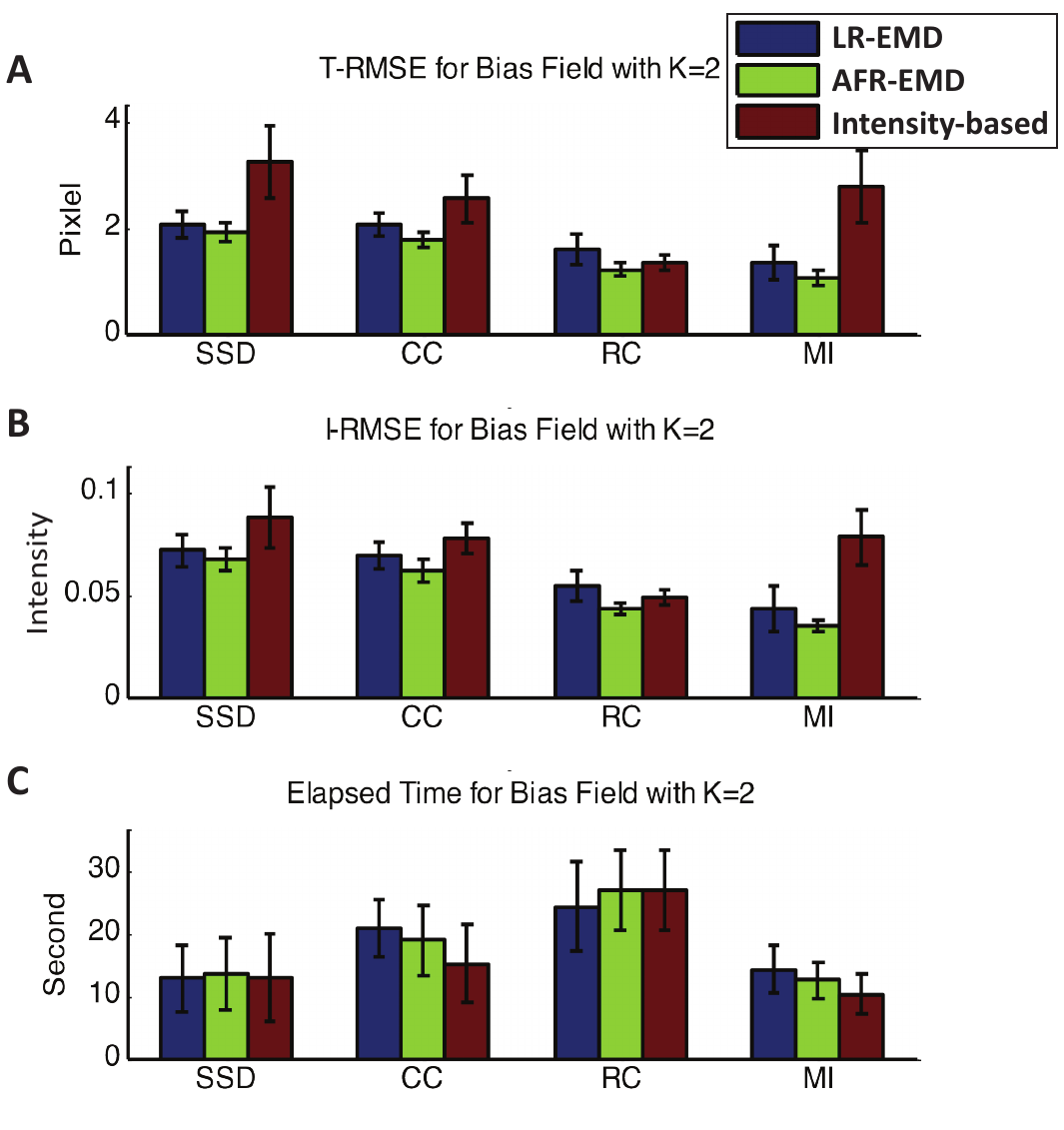}
  \captionof{figure}{Registration performance in the presence of two Gaussian kernels. Mean and variance of T-RMSE and I-RMSE for images in BrainWeb dataset are shown in top and middle panels. Bottom panel compares the running time for all methods.}
  \label{fig:4-16}
\end{minipage}
\end{figure*}

\subsubsection{Convergence in the presence of bias field}

Next, we investigate the registration of the MR images in the presence of the bias field noise. This setting is particularly of interest to us because intensity-based algorithms often fail to converge when images are distorted by the bias field noise. Even converging algorithms have less accuracy compared to the registration of the same image without any bias field noise. To study the convergence of registration algorithms when images are noisy, we manually added Gaussian functions to both floating and reference images according to Equation \ref{gaussian}. Then, we performed image registration using LR-EMD, AFR-EMD and the intensity-based benchmark algorithm. An algorithm is considered converged when the T-RMSE error is lower than 4 pixels. For each image in the BrainWeb dataset, we repeated the registration process for 15 random selection of the Gaussian kernel centers. The convergence rate is then defined as the number of converged algorithms divided by 15 (times 100 to get the percentage).

Figure \ref{fig:4-11} shows the convergence rate for each of the four similarity measures (SSD, CC, RC and MI). The number of Gaussian kernels are varied between $K=0$ (no bias field) to $K=4$ (Four Gaussian kernels in each of the floating and reference images). For SSD, CC an MI similarity measures, adding one Gaussian function remarkably degrades the convergence rate for the intensity based algorithm (red curves in Figure \ref{fig:4-11}). In particular, SSD never converged with one Gaussian kernel. This confirms previous observations that the intensity-based registration is not robust to the bias field noise. RC is the most robust similarity measure for the intensity-based registration with 80\% convergence rate in the presence of one Gaussian kernel and  100\% for two to four Gaussian kernels. Moving to our proposed EMD-based algorithms, we observe that both LR-EMD and AFR-EMD improve the convergence rate for almost all of the cases. Using MI or CC as the similarity measure, both LR-EMD and AFR-EMD converge 100\% of the times while intensity-based registration converges in 20\% to 60\% of the times for MI and 25\% to 80\% of the times for CC. For SSD, both LR-EMD and AFR-EMD have higher convergence rate compared to benchmark registration for one to three Gaussian kernels. When four Gaussian kernels are present, LR-EMD has lower convergence rate compared to other two algorithms. This is primarily because of the low SNR in images. Overall, our results suggest that both EMD-based algorithms as well as the intensity-based registration with the RC similarity measure have higher convergence rate compared to other similarity measures.

\subsubsection{Registration performance in the presence of bias field}

In this section, we study the performance of the registration algorithms across the experiments that have converged. First, we explore the case that each of the reference and floating images are corrupted by one Gaussian function ($K=1$). Top two panels in Figure \ref{fig:4-15} show the T-RMSE and I-RMSE for the benchmark intensity-based algorithm as well as LR-EMD and AFR-EMD. For the SSD similarity measure, intensity-based registration has never converged, therefore, no corresponding error value is reported. For the MI similarity measure, both LR-EMD and AFR-EMD are remarkably more accurate compared to the intensity-based registration (50\% lower T-RMSE and 40\% lower I-RMSE). MI is a particularly interesting case because registration algorithms based on MI generally achieve higher accuracy compared to other three similarity measures and are often faster (Figure \ref{fig:4-15} bottom panel). For the CC similarity measure, both EMD-based registrations have $\sim$20\% lower average error rate. Using RC as the similarity measure, LR-EMD has slightly higher average error rate compared to two other methods. Bottom panel in Figure \ref{fig:4-15} presents a comparison of the running time between benchmark and two proposed algorithms. The intensity-based registration using CC similarity measure converges $\sim$50\% faster compared to our two methods. All three algorithms have a similar running time when using MI as the similarity measure. Overall, both LR-EMD and AFR-EMD with the MI similarity measure achieve the highest accuracy with the lowest running time and are best performing algorithms in the presence of one Gaussian kernel in the images.

\begin{figure*}[t]
  \centering
  \includegraphics[width=1\textwidth]{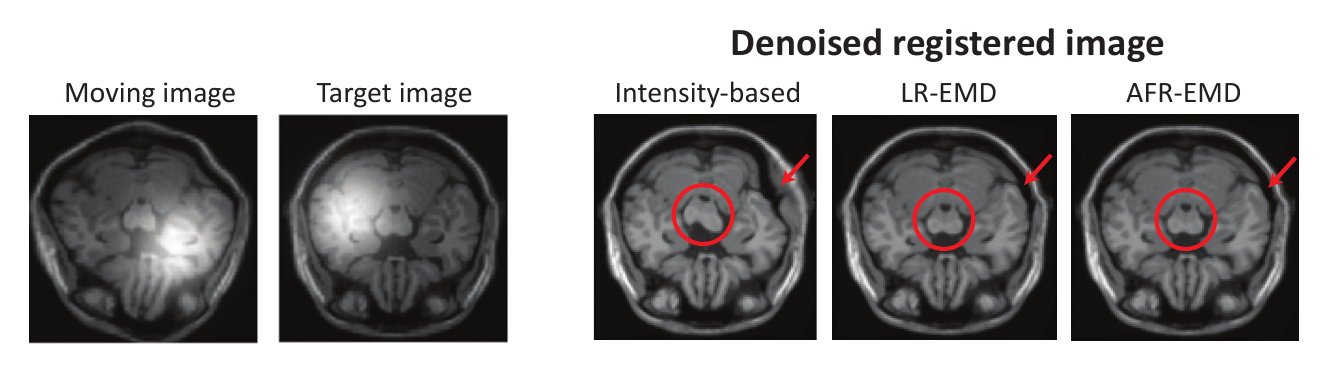}
  \caption{Target and reference images from BrainWeb MR dataset and registered images using our two methods and intensity-based hierarchical algorithm. MI is used as the similarity measure. Visually, both LR-EMD and AFR-EMD improve the registration accuracy in the center of image (red circle) and at the edge (red arrow).}
  \label{fig:brainweb}
\end{figure*}

To present a more comprehensive comparison between our EMD-based registration algorithms and the benchmark method, one set of registered images are visualized in Figure \ref{fig:brainweb}. The image is randomly chosen from the BrainWeb dataset, then geometrically distorted using a random perturbation, and finally contaminated with $K=1$ Gaussian function. In this case, MI is used as the similarity measure. Two images in the left panel of Figure \ref{fig:brainweb} are the moving and the target images, both contaminated with bias field noise. The denoised registered images are shown in the right for our two proposed algorithms and the intensity-based benchmark algorithm. Visually, both LR-EMD and AFR-EMD perform a more accurate registration compared to the intensity-based method for this image. The improvement is visible both in the cerebral peduncle (red circle) and within the top right area of the skull (red arrow). 

Figure \ref{fig:4-16} shows the accuracy and running time for registrations in the presence of two Gaussian kernels in each of the floating and reference images ($K=2$ case). Two Gaussian kernels simulates a more intense bias field noise. Using any of the SSD, CC or MI similarity measures, EMD-based registrations achieve higher accuracy both in T-RMSE and I-RMSE. For MI, EMD-based registrations have $\sim$60\% lower T-RMSE and $\sim$50\% lower I-RMSE compared to the benchmark method. For all similarity measures, LR-EMD has slightly higher accuracy compared to AFR-EMD. 

Another observation that is worth noting is that for the intensity-based benchmark algorithm, RC similarity measure has at least $\sim$50\% higher accuracy compared to any other similarity measures. This is consistent with observations in the previous studies on the robustness of RC to bias filed noise \cite{myronenko2010intensity}. However, the running speed is at least $\sim$60\% higher compared to other similarity measures. That is, there is a trade off between accuracy and running speed for the RC similarity measure. Interestingly, this is not the case for the EMD-based algorithms when using MI similarity measure. Both LR-EMD and AFR-EMD achieve the highest accuracy and lowest running time compared to all other algorithms.  

\begin{figure*}[!t]
\centering
\begin{minipage}{.48\textwidth}
  \centering
  \includegraphics[width=1\textwidth]{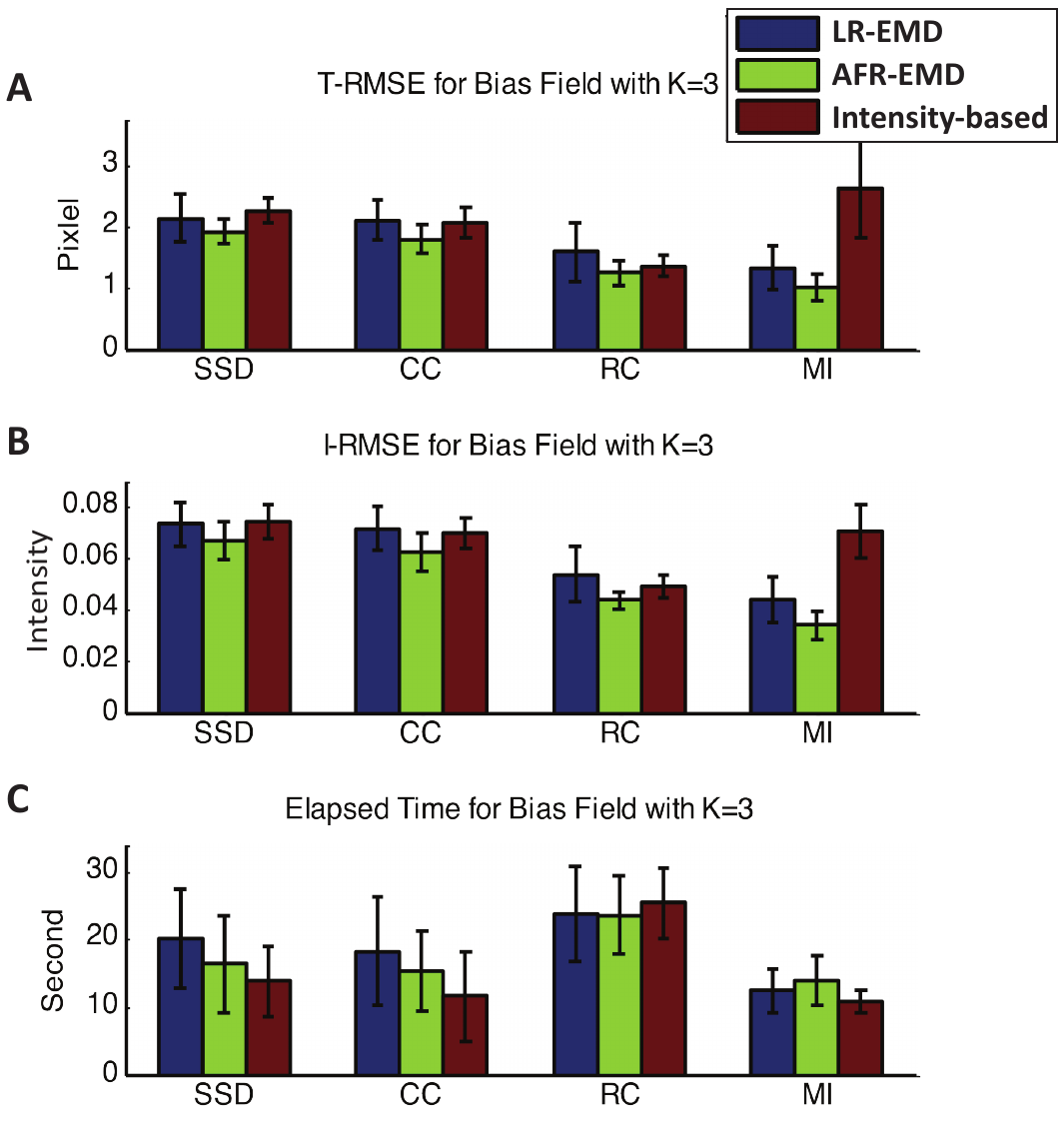}
  \caption{Registration performance in the presence of three Gaussian kernels. Mean and variance of T-RMSE and I-RMSE for images in BrainWeb dataset are shown in top and middle panels. Bottom panel compares the running time for all methods.}
  \label{fig:4-17}
\end{minipage}%
\hfill
\begin{minipage}{.48\textwidth}
  \centering
  \includegraphics[width=1\textwidth]{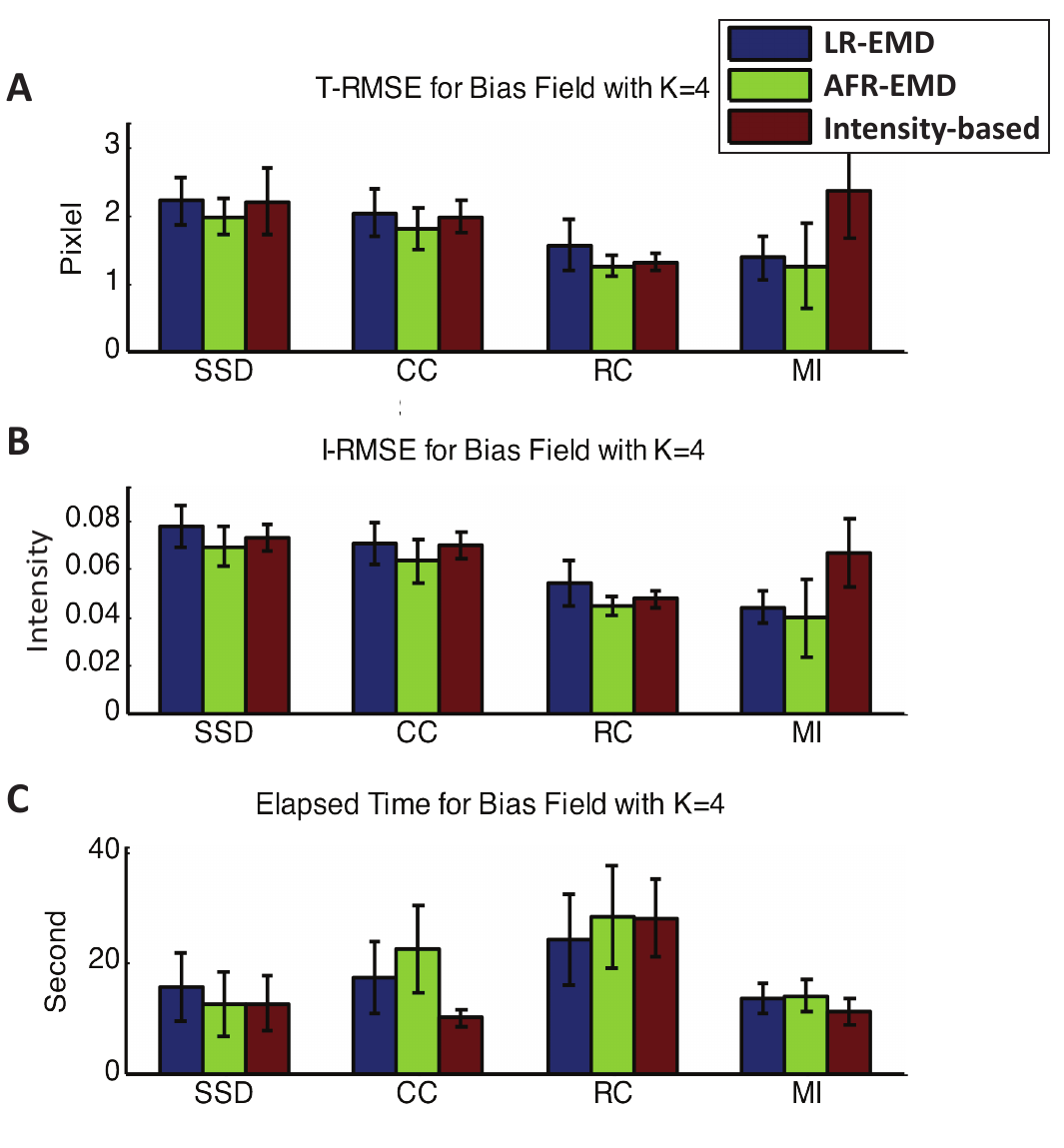}
  \caption{Registration performance in the presence of four Gaussian kernels. Mean and variance of T-RMSE and I-RMSE for images in BrainWeb dataset are shown in top and middle panels. Bottom panel compares the running time for all methods.}
  \label{fig:4-18}
\end{minipage}
\end{figure*}

 We have also determined the registration performance in the presence of three and four Gaussian kernels and reported the T-RMSE, I-RMSE and the running time in Figures \ref{fig:4-17} and \ref{fig:4-18}. Using SSD, CC or RC as the similarity measure, EMD-based methods have approximately similar performance to the intensity-based registration. When using MI as the similarity measure, EMD-based registrations achieves 40\% to 65\% higher accuracy both in T-RMSE and I-RMSE compared to the intensity-based registration. Similar to the previous cases, MI attains the most accurate and the fastest registration compared to other three similarity measures. Overall. our results suggest that MI is the most suited similarity measure to the EMD-based registration techniques. 

A summary of the quantitative performance measures are presented in Table \ref{table:mse16by16}. T-RMSE, I-RMSE, and the convergence percentages reported in this table are the average values across all the bias field noise conditions ($0 \le K \le 4$). Using MI as similarity measure, AFR-EMD  achieves 42\% lower error rate in intensity and 52\% lower error rate in transformation compared to intensity-based hierarchical registration. AFR-EMD with MI achieves 27\% lower error rate in intensity and 21\% lower error rate for transformation compared to intensity-based registration with RC similarity measure (which is robust to bias field noise). For LR-EMD, the error rate is 32\% lower for intensity and 41\% lower for transformation when using MI similarity measure.

 \begin{table*}[t]
  \caption{Summary of the quantitative performance measures}
  \centering
  \scalebox{0.9}{
  \begin{tabular}{c c c c c}
  \hline\hline
  Similarity measure & Method & Convergence percentage & T-RMSE & I-RMSE  \\  
  \hline
  \hline

\multirow{3}{*}{SSD} & Intensity-based  & 61.33\% & 2.393 $\pm$ 0.375  & 0.075 $\pm$ 0.008\\ 
 & LR-EMD & 80\%  & 2.088 $\pm$ 0.309  & 0.073 $\pm$ 0.008\\ 
 & AFR-EMD & \textbf{90\%} & \textbf{1.927 $\pm$ 0.206}  & \textbf{0.067 $\pm$ 0.007}\\ 
 
 \hline
 
\multirow{3}{*}{CC} & Intensity-based  & 76\% & 2.167 $\pm$ 0.341  & 0.069 $\pm$ 0.007\\ 
 & LR-EMD & \textbf{100\%}  & 2.032 $\pm$ 0.323  & 0.069 $\pm$ 0.008\\ 
 & AFR-EMD & \textbf{100\%} & \textbf{1.777 $\pm$ 0.213}  & \textbf{0.062 $\pm$ 0.007}\\ 
 
 \hline

\multirow{3}{*}{RC} & Intensity-based  & 96\% & 1.331 $\pm$ 0.142  & 0.048 $\pm$ 0.003\\ 
 & LR-EMD & 97.33\%  & 1.544 $\pm$ 0.357  & 0.053 $\pm$ 0.009\\ 
 & AFR-EMD & \textbf{100\%} & \textbf{1.217 $\pm$ 0.153}  & \textbf{0.043 $\pm$ 0.003}\\ 
 
 \hline
 
\multirow{3}{*}{MI} & Intensity-based  & 53\% & 2.205 $\pm$ 0.516  & 0.062 $\pm$ 0.009\\ 
 & LR-EMD & \textbf{100\%}  & 1.299 $\pm$ 0.325  & 0.042 $\pm$ 0.008\\ 
 & AFR-EMD & \textbf{100\%} & \textbf{1.054 $\pm$ 0.266}  & \textbf{0.035 $\pm$ 0.007}\\ 
 
 \hline

  \end{tabular}
  }
  \label{table:mse16by16}
\end{table*}

\subsection{Retina images}
We have argued that the proposed EMD-based approach is successful in the registration of two images in the presence of spatially varying intensity distortion. This feature is not limited to MR images. Here, we evaluate the proposed approach on retina images and in the presence of spatially varying intensity distortion. Retina images are frequently used in the diagnosis of various ophthalmological disorders. These images are often taken at different times. Hence image registration has a crucial role in accurate comparison of the longitudinal retina images. However, these images are often distorted via a non-uniform background or spatially varying intensity noise. Previous studies have explored the application of vascular structures and landmarks in the registration of these noisy images \cite{ghassabi2013efficient, stewart2003dual}. However, the registration performance still remains far from perfect. Here, we investigate the application of LR-EMD and AFR-EMD in the registration of retina images.

\begin{figure*}[!t]
  \centering
  \includegraphics[width=.9\textwidth]{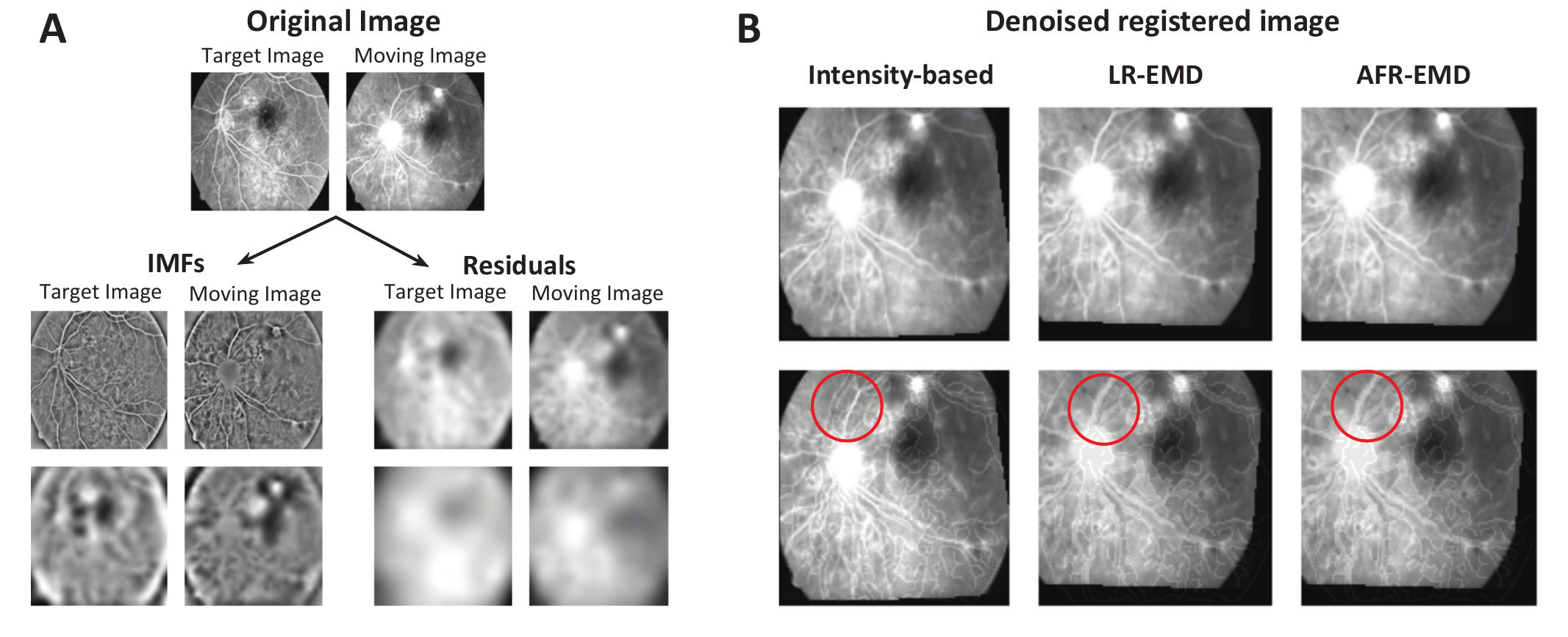}
  \caption{Registration of Retina images taken 2 years apart. \textbf{A.} The set of IMFs and residuals for two levels of EMD are shown for both target and moving images. \textbf{B.} Registered images using our two methods and intensity-based hierarchical algorithm. The second row shows the composite view through contour overlap after the registration.}
  \label{fig:retina}
\end{figure*}

To evaluate the performance of our proposed methods, we used two retina images taken two years apart (Fig. \ref{fig:retina}.A top panel) \cite{zana1998region}. Both images were distorted via strong spatially-varying intensity noise. Figure \ref{fig:retina}.A illustrates two levels of the IMFs and residuals for each image. The residuals contain the significant part of the spatially varying noise and the IMFs contain denoised patterns. Therefore, similar to brain BR images, the EMD-based features have decomposed the signal and noise into two separate components. Figure \ref{fig:retina}.B demonstrates the performance of registration based on LR-EMD, AFR-EMD as well as the benchmark intensity-based algorithm for these retina images. Here, SSD is used as the similarity measure. From the figure, both LR-EMD, AFR-EMD visually outperform the intensity based registration. This is demonstrated via the composite view through contour overlap after the registration in Figure \ref{fig:retina}.B. In particular, the red circle shows the vascular pattern for which the intensity based approach has failed to accurately register. Both LR-EMD and AFR-EMD are successful in registering this vascular pattern. This visual inspection suggest that both of the proposed methods are robust against the spatially varying intensity distortion in the retina images.

\section{Discussion and future works}

Throughout this paper, we presented evidences on the advantage of using empirical mode decomposition in the registration of single modal MR images. However, we believe that EMD has potentially a wide variety of applications in multi-modal image registration, image fusion, and image denoising. Particularly, multi-modal image registration has a great potential for more investigations because EMD-based features are relatively robust to the modality. IMFs are made of normalized minimum and maximum envelopes of intensity. Therefore, the effect of modality-related variabilities are much less in IMFS. The idea of using EMD for multi-modal registration has been previously discussed \cite{jinsha2009multimodal}, however, a more careful investigation of the applications of EMD in multi-modal image registration such as CT-MR image registration is still necessary.

In this paper, we did not consider the problem of parametrization for the transformation. We have used free form deformation to formulate the transformation, however, EMD embedded in other transformation techniques needs further investigation. In particular, the application of nonlinear transforms such as neural networks \cite{specht1991general} and random forest \cite{breiman2001random} in EMD-based registration needs extensive study. We believe that compressed non-linear networks \cite{abbasi2017structural} together with EMD could achieve fast and accurate registration in the presence of bias field noise. One advantage of these models is that the transform is more interpretable \cite{abbasi2017interpreting, vellido2012making}. Other nonlinear modular techniques such as Hammerstein Wiener model have shown impressive performance for applications such as nonlinear mapping in biomedical signal processing \cite{abbasi2011estimation, hunter1986identification} and computational Neuroscience \cite{abbasi2012hammerstein}. Employing these models as a core transformation technique between two images and possible EMD-based boost remain for future work.

Finally, we believe that EMD-based registration techniques generalize well to other registration application. Remote sensing and satellite images, other medical applications such as retina and angiography image registration and computer vision applications such as stereo vision are a few examples. An extensive study on the performance of EMD-based registration for these applications is necessary in the future.

\section*{Data availability}
All the data used in this manuscript is available online \cite{BrainWeb}.

\section*{Competing interests}
The authors declare no competing interests.

\section*{Author contributions statement}
R.A. and E.F. conceived the experiments, R.A. and A.G. conducted the experiment(s). All authors analyzed the results. R.A. wrote the manuscript with contributions from A.G and E.F. All authors reviewed the manuscript.





\end{document}